\documentclass[final,12pt]{clear2025}

\usepackage{amsmath,amsfonts,bm}

\def\eqref#1{equation~\ref{#1}}

\def\1{\bm{1}}

\def\rg{{\textnormal{g}}}

\def\rvx{{\mathbf{x}}}

\def\mA{{\bm{A}}}

\def\mC{{\bm{C}}}

\def\mE{{\bm{E}}}

\def\mI{{\bm{I}}}

\def\mM{{\bm{M}}}
\def\mN{{\bm{N}}}

\def\mX{{\bm{X}}}

\DeclareMathAlphabet{\mathsfit}{\encodingdefault}{\sfdefault}{m}{sl}
\SetMathAlphabet{\mathsfit}{bold}{\encodingdefault}{\sfdefault}{bx}{n}

\newcommand{\R}{\mathbb{R}}

\usepackage{times}
\usepackage{hyperref}
\usepackage{url}
\usepackage{adjustbox}
\usepackage{cleveref}
\usepackage{booktabs}
\usepackage[font=small, skip=2pt]{caption}
\usepackage[symbol]{footmisc}

\title[The CausalBench challenge]{The CausalBench challenge: A machine learning contest for gene network inference from single-cell perturbation data}

\clearauthor{%
{\normalfont \large Organizers:}\\
\\
\textbf{Mathieu Chevalley}$^{5,7}$ \quad \textbf{Jacob Sackett-Sanders}$^7$ \quad \textbf{Yusuf Roohani}$^{7,8}$ \quad \textbf{Pascal Notin}$^{9, 10}$  \\ \textbf{Arash Mehrjou}$^{7\dagger}$ \quad \textbf{Patrick Schwab}$^{7\dagger}$
\\
\\
{\normalfont \large Participants:}\\
\\
\textbf{Artemy Bakulin}$^{2}$ \quad 
\textbf{Dariusz Brzezinski}$^{1}$ \quad 
\textbf{Kaiwen Deng}$^3$ \quad 
\textbf{Yuanfang Guan}$^3$ \quad \\
\textbf{Justin Hong}$^4$ \quad
\textbf{Michael Ibrahim}$^1$ \quad 
\textbf{Wojciech Kotlowski}$^1$ \quad 
\textbf{Marcin Kowiel}$^6$ \quad \\
\textbf{Panagiotis Misiakos}$^5$ \quad
\textbf{Achille Nazaret}$^4$ \quad 
\textbf{Markus Püschel}$^5$ \quad 
\textbf{Chris Wendler}$^5$
\\
\\
{\normalfont $^1$Poznan University of Technology \quad $^2$Lomonosov Moscow State University
$^3$University of Michigan \quad $^4$Columbia University \quad  $^5$ETH Zürich \quad $^6$Ryvu Therapeutics \quad $^7$GSK.ai \quad
$^8$Stanford University \quad
$^9$Harvard Medical School
$^{10}$University of Oxford}
}

\begin{document}

\maketitle

\begin{abstract}
\looseness=-1 In drug discovery, mapping interactions between genes within cellular systems is a crucial early step. Such maps are not only foundational for understanding the molecular mechanisms underlying disease biology but also pivotal for formulating hypotheses about potential targets for new medicines. Recognizing the need to elevate the construction of these gene-gene interaction networks, especially from large-scale, real-world datasets of perturbed single cells, the CausalBench Challenge was initiated. This challenge aimed to inspire the machine learning community to enhance state-of-the-art methods, emphasizing better utilization of expansive genetic perturbation data. Using the framework provided by the CausalBench benchmark, participants were tasked with refining the current methodologies or proposing new ones. This report provides an analysis and summary of the methods submitted during the challenge to give a partial image of the state of the art at the time of the challenge. Notably, the winning solutions significantly improved performance compared to previous baselines, establishing a new state of the art for this critical task in biology and medicine.
\end{abstract}

\section{Introduction}

\footnotetext[0]{$^\dagger$ Jointly advising}

\looseness=-1 Causal gene interaction networks, visualized as directed graphs, visually elucidate the complex interactions of genes within a cell, revealing cell specific cellular function and regulation. These networks are not only instrumental in differentiating normal and aberrant cellular processes but also illuminate pathways perturbed in various diseases. By studying these networks, we gain insights that can guide the identification of potential drug targets \citep{nelson2015support,yu2004advances,chai2014review,akers2021gene,hu2020integration}. More crucially, deriving these interactions allows researchers to anticipate potential unsafe downstream effects of targets, ensuring that therapeutic interventions are both effective and safe. 
The ability to accurately infer these networks from experimental data, therefore, holds profound implications for biological research and medicine.

\looseness=-1 Recent strides in single-cell transcriptomics have ushered in tools that directly target and suppress gene expression \citep{dixit2016perturb,datlinger2017pooled,datlinger2021ultra}. Given that those perturbations represent actual interventions in the cell system, the experimental data produced holds great promise to reveal the underlying causal mechanisms dictating cell function \citep{pearl2009causality}. Causal discovery methods that leverage interventional data by explicitly modeling the interventional signal are thus primed candidate approaches to infer gene interaction networks. CausalBench, introduced by \citet{chevalley2022causalbench}, was envisaged to evaluate network inference methodologies using single-cell perturbation gene expression data. CausalBench operates on two recent expansive public CRISPR-based perturbation datasets \citep{replogle2022mapping} of unprecedented scale and introduces novel evaluation metrics that are biologically relevant. However, early evaluations revealed a stark surprise. Notably, the performance of leading-edge inference methods plateaued despite the increased perturbation data. Even more surprisingly, methods incorporating interventional data were observed not to outperform methods that did not utilize such data, indicating only marginal utility from interventional signals opposing the common belief that interventional data is all you need to resolve the ambiguity in causal inference tasks with existing methods \citep{hauser2012characterization,eberhardt2006n,yang2018characterizing}.

\looseness=-1 These observations suggest a large gap for interventional causal discovery methods between their reported performance on synthetic and real-world data. Indeed, many researchers have prioritized evaluation on synthetic datasets due to the allure of having an available ground-truth, which seemingly facilitates method comparison. However, as recently highlighted by benchmark evaluation such as CausalBench, the reported performance on synthetic data often does not translate well to real-world settings. This discrepancy might deter practitioners due to the unpredictability of success in real-world settings. And even more disconcerting is the prospect of blind reliance on synthetic data results. Given the actual poor performance of those models in real-world setting, this could culminate in large misplaced resources in terms of time, money, and missed opportunities. 

\looseness=-1 To bridge this gap and spearhead advancements in gene interaction network inference, we organized a machine learning community challenge named the CausalBench Challenge (CBC2023) \url{https://www.gsk.ai/causalbench-challenge/}. 
The aim behind the challenge was to lower the bar of engagement and inspire the machine learning community to work on this critical task and elevate the current state of the art.
In this report, we delineate how the methods proposed by the participants have remarkably advanced the state of the art performance on this essential endeavor. This epitomizes the effect of community scientific competition such as CBC2023 on precipitating scientific breakthroughs. Within a limited timeframe, we witnessed advancements that might otherwise have required a significantly longer time frame to conduct. Given the significance of gene network inference in pivotal domains like drug discovery, any expedited progress can yield profound societal benefits. 

\looseness=-1 To facilitate the application of the new methods as well as future research, the method implementations have been open sourced. Furthermore, participants have provided detailed reports on each method, available at the challenge's OpenReview venue \url{https://openreview.net/group?id=GSK.ai/2023/CBC}.

To summarize, in this report:

\begin{itemize}
    \item We demonstrate that Machine Learning contests can be effective tools to engage the community on impactful problems. Notably, many participants had no prior experience in gene network inference or even in biology.
    \item We describe how the participants were able to focus on technical innovation. This was achieved through the establishment of clear, targeted goals, suitable computing infrastructure, and the use of a curated benchmark as a development platform.
    \item We conduct a thorough analysis of the best solutions to establish a new state of the art in the task of gene network inference. By providing a comprehensive overview of key ideas, discussing remaining limitations, and making the method implementations open-source, we hope to pave the way for further major advancements in this vital area.
\end{itemize}

\section{Benchmark Setup}

\looseness=-1 CausalBench \citep{chevalley2022causalbench} is a comprehensive benchmarking suite designed to evaluate network inference methods for gene regulatory networks (GRNs) from real-world single-cell perturbation data. The benchmark is built on large-scale, high-dimensional datasets obtained from single-cell RNA sequencing (scRNA-seq) experiments \citep{replogle2022mapping}, where genes are perturbed using CRISPR technology to observe gene expression changes. This setup provides a more realistic evaluation of causal inference methods compared to traditional synthetic benchmarks \citep{gentzel2019case}.

\looseness=-1 \paragraph{Datasets} The CausalBench benchmark relies on two large-scale perturbational datasets, each consisting of gene expression data from individual cells in two cell lines: RPE1 (retinal pigment epithelial cells) and K562 (human chronic myelogenous leukemia cells). These datasets contain over 200,000 interventional data points, where each perturbation corresponds to the knockout or knockdown of a specific gene using CRISPR technology \citep{dixit2016perturb,datlinger2017pooled,datlinger2021ultra}. Observational data is collected without interventions, while the interventional data includes targeted gene perturbations. 


\looseness=-1 \paragraph{Task} The primary goal of the task is to infer the true causal graph \( \mathcal{G} \) between a set of variables from empirical data. In the context of gene regulatory networks (GRNs), these variables represent genes, and the causal graph encodes the functional dependencies between them. Each node in the graph corresponds to a gene \( g_i \), and a directed edge from node \( g_i \) to node \( g_j \) signifies a causal influence of gene \( g_i \) on gene \( g_j \). The challenge lies in reconstructing this true graph \( \mathcal{G} \) from observed data, typically involving both observational data (where no interventions are applied) and interventional data (where specific genes are perturbed).

\looseness=-1 We define several notations that will be used throughout the paper. \( V \) represents the set of all genes in the network. The true causal graph is denoted \( \mathcal{G} = (V, E) \), where \( E \) is the set of directed edges between genes, representing causal relationships. For each gene \( g_i \), \( \rvx_{g_i} \in \mathbb{R}^n \) refers to the observed data, with \( n \) being the number of samples (cells). The full observed data for all genes is represented by \( \rvx = (\rvx_{g_1}, \rvx_{g_2}, \dots, \rvx_{g_m}) \), where \( m \) denotes the total number of genes in the dataset. \( P^{\emptyset}(\rvx_{g_j}) \) is the marginal probability distribution of gene \( g_j \) under no interventions (observational distribution), representing the natural state of gene expression without perturbation, and \( P^{\sigma(\rvx_{g_i})}(\rvx_{g_j}) \) denotes the probability distribution of gene \( g_j \) under the intervention on gene \( g_i \), where \( \sigma(\rvx_{g_i}) \) represents an intervention on gene \( g_i \).

\looseness=-1 The goal is to infer the graph \( \mathcal{G} \) from the data \( \rvx \), ideally reconstructing the correct edges that reflect true causal relationships in the biological system. We denote a predicted graph $G_p$. To accomplish this, both observational data \( \rvx \) and interventional data \( \rvx_\sigma \) are used. Observational data provides information about the natural relationships between genes, while interventional data allows us to better identify causal effects by perturbing specific genes. The challenge in causal discovery lies in utilizing both types of data to uncover the underlying causal graph, while addressing the complexities and limitations inherent in real-world biological systems.

\looseness=-1 \paragraph{Metrics}
CausalBench incorporates two evaluation approaches: a biologically-motivated evaluation and a statistical evaluation. The biologically-motivated evaluation uses known biological databases, such as CORUM \citep{giurgiu2019corum} and STRING \citep{von2005string, snel2000string, von2007string, jensen2009string, mering2003string, szklarczyk2010string, szklarczyk2015string, szklarczyk2016string, szklarczyk2019string, szklarczyk2021string, franceschini2012string, franceschini2016svd}, to compare the predicted gene-gene interactions with established interactions. This allows for precision and recall measures based on known relationships, but it is limited by lack of cell-specific data, and may not cover all causal relationships in the system.

For the CausalBench challenge, however, we focus on the statistical evaluation, which avoids relying on prior knowledge. This approach leverages the mean Wasserstein distance \citep{ramdas2017wasserstein} and false omission rate (FOR) metrics proposed in CausalBench, which are computed on a held-out test dataset. The Mean Wasserstein Distance metric measures the divergence between the observational and interventional distributions of gene expressions. A higher Wasserstein distance indicates stronger causal effects for the predicted edges. Formally, it is computed as:

\begin{equation}
\text{MeanWasserstein} = \frac{1}{| G_p |} \sum_{(i, j) \in  G_p } W_1(P^{\sigma(X_{g_i})}(\rvx_{g_j}), P^{\emptyset}(\rvx_{g_j}))
\end{equation}

where $P^{\sigma(X_{g_i})}(\rvx_{g_j})$ is the distribution fo gene $g_j$ under intervention of gene $g_i$, and $P^{\emptyset}(\rvx_{g_j})$ is marginal distribution of $g_j$ under no interventions (observational distribution). 

The FOR metric evaluates the rate at which true causal interactions are missed by the model. It helps evaluate the balance between precision and recall in the predicted network. Intuitively, there is a trade-off between maximizing the mean Wasserstein distance and minimizing the FOR. Formally, the FOR is defined as  

\begin{equation}
    \text{FOR} = \frac{\text{False Negatives}}{\text{False Negatives} + \text{True Negatives}}.
\end{equation}

This ratio is estimated by finding false negatives with a two-sided Mann–Whitney U rank test \citep{mann1947test, Bucchianico1999-ul} on sampled \emph{negative} pairs. Negative pairs of gene $(g_i, g_j)$ are variable for which there is no directed path in the predicted causal graph $G_p$. The statistical test is performed on $P^{\sigma(X_{g_i})}(\rvx_{g_j})$ and  $P^{\emptyset}(\rvx_{g_j})$. The focus on statistical evaluation allows for an objective and empirical assessment of the methods, particularly for real-world interventional data, where true causal graphs are not available.

\paragraph{The CausalBench Challenge}
\looseness=-1 In the original CausalBench study, several limitations of existing methods were identified, which motivated the design of the CausalBench Challenge. Two key limitations were observed. First, many existing methods struggled to scale to large gene networks, especially in the context of high-dimensional single-cell data, which is essential for gene regulatory network inference at scale. Second, despite the availability of rich interventional data, many methods did not fully leverage this information, leading to suboptimal performance when compared to methods trained solely on observational data. The CausalBench Challenge was specifically designed to address these limitations by incentivizing participants to develop methods that could better scale to large networks and more effectively utilize interventional data. Details about the challenge design are described in \cref{app:design}. By focusing on the statistical evaluation metrics, the challenge sought to push the boundaries of current methods and improve their ability to handle the complex, high-dimensional nature of the datasets. The results of the challenge demonstrated significant improvements over previous methods, particularly in terms of scalability and the use of interventional information. These advancements highlight the importance of improving both the scalability and the ability to integrate interventional data in causal network inference, which are key factors for advancing the field.

\paragraph{Limitations}
\looseness=-1 While CausalBench offers an important advancement in benchmarking causal discovery methods, it also has some limitations. One key limitation is the difficulty in distinguishing between direct and indirect causal relationships using the statistical metrics. The mean Wasserstein distance, while useful for measuring the strength of causal effects, does not differentiate between direct and indirect relationships, which can impact the interpretation of the results. Additionally, since there is no true causal graph, the evaluation provides relative assessments of method performance based on known biological interactions, which may not encompass all true causal relationships in the system.  These limitations, particularly the inability of the statistical metrics to distinguish direct from indirect causal effects, must be taken into account when analyzing the evaluated performance. The metrics in CausalBench inherently test for a different property compared to traditional metrics like Structural Hamming Distance (SHD) and Structural Intervention Distance (SID), which require a known ground truth. The metrics of CausalBench were developed to address the absence of ground truth in this real-world setting. However, the statistical approach may offer a more nuanced comparison of methods by weighting predictions and errors according to the strength of causal effects, rather than treating all edge errors equally. More research is needed to better understand how these metrics influence the interpretation of method performance and how they compare to traditional benchmarks. 

\section{Methods}


We here summarize the theoretical ideas behind the best-submitted solutions. More details for each method can be found in the reports submitted to the challenge public venue hosted on OpenReview \url{https://openreview.net/group?id=GSK.ai/2023/CBC}.

\paragraph{BetterBoost - Inference of Gene Regulatory Networks with Perturbation Data} \citep{nazaret2023betterboost}
BetterBoost builds upon GRNBoost to harness the power of interventional data obtained from Perturb-seq experiments. While GRNBoost ranks directed interactions between genes based on their predictiveness, BetterBoost takes it a step further by incorporating perturbation data and its underlying causal structure to create a new score for ranking interactions.

In GRNBoost, each directed interaction from a gene $g_i$ to a gene $g_j$ is assigned a score $G_{i,j}$, indicating the predictive capability of gene $g_i$ for the target gene $g_j$. This score is computed using boosted trees, which predict the expression of gene $g_j$ based on all other genes. A feature-importance metric then scores each potential parent gene $g_i \neq g_j$. However, this approach is purely prediction-based and lacks the incorporation of causal relationships. With BetterBoost, we integrate perturbation data and its underlying causal structure to establish a new score for ranking interactions, denoted as $H_{i,j}$. 

Intuitively, if a candidate gene is a parent of the target, it should be a good predictor for the target, as GRNBoost assumes. But with additional labeled interventional data, one can attempt to identify the true causal parents of a target gene $g_j$ by looking at the effects of interventions on the candidate parents for $g_i$. In particular, for a true causal parent $g_i$, we expect that when $g_i$ is knocked down, there will be a statistically significant shift in the distribution of observed Unique Molecular Identifiers (UMIs) of gene $g_j$  between observational and interventional data. Since we hold no priors on the nature of causal effects, we choose to use the Kolmogorov-Smirnov (KS) test to test these distributional shifts between observational and interventional data. Furthermore, we correct the p-values for multiple testing with the Benjamini-Hochberg procedure.

To formulate the new score utilized by BetterBoost in ranking the impact of gene $g_i$ on gene $g_j$, we compute the predictive score $G_{i,j}$ obtained from GRNBoost and the Benjamini-Hochberg corrected p-value $p_{i,j}$ from the KS test, which measures the impact of knocking down gene $g_i$ on gene $g_j$. In cases where no perturbation data for gene $g_i$ is available, we set all p-values $p_{i, *}$ to $0.05$. This choice aims to neither strongly accept nor reject hypotheses regarding these interactions. Consequently, we define the score $B_{i,j} = (- p_{i,j}, G_{i,j})$, which we sort in descending order (using lexicographic order). It should be noted that alternative methods of combining $p_{i,j}$ and $G_{i,j}$ could be explored in future research.

When a desired number of edges, denoted as $K$, is specified, BetterBoost returns the top $K'$ candidate edges with the highest $H$ score and acceptable p-values (where $K'$ is $K$ if there is enough acceptable p-values $\leq 0.05$, or is less otherwise). These candidate edges will possess the smallest p-values for the KS test up to 0.05 (included), including gene pairs without available interventional data and hence with default p-values of 0.05. By setting the p-values of these gene pairs to $0.05$, the ranking primarily favors edges with small p-values (derived from combined interventional and observational data), followed by edges with the highest GRNBoost scores $G_{i,j}$ (obtained solely from observational data). Typically, this results in more of the final edges being chosen by p-value than by GRNBoost score as more labeled interventional data becomes available.

\paragraph{Guanlab: A Supervised LightGBM-Based Approach} \citep{deng2023supervised}
The basic idea of our LightGBM-based approach was transforming the task of detecting gene pairs with causal relationships into a supervised learning problem. Given the gene pairs that were more confident to have causalities, a supervised-learning model should be able to retrieve the expression properties that determined such relationships. We chose the LightGBM as the base learner, a tree-based learning algorithm that was widely used in various classification tasks, and build a new dataset from the observational and interventional data for training and evaluation. A major challenge in constructing this dataset was to determine the positive samples. We used the absolute correlations between the expression of two different genes to quantify the causality strengths. A gene pair with a stronger negative or positive expression correlation was more likely to be affected by each other. And the correlation values were also available when there was no interventional data.

We calculated the Pearson correlations for all directed gene pairs $\langle\rg_{i}, \rg_{j}\rangle$ which indicated $\rg_{i}$ affect $\rg_{j}$. $\rg_{i}$ and $\rg_{j}$ were genes with expression measurements from the columns of the expression matrix $\mE$ and $i \neq j$. Let the $i$, $j$ be the column indexes of $\rg_{i}$ and $\rg_{j}$, $i^{\prime}$, $j^{\prime}$ be the row indexes for all records intervened by $\rg_{i}$ and $\rg_{j}$, and $n$ be the row indexes of all observational records, we first retrieved the intervened expression data: $\mE_{i^{\prime}, i}$ and $\mE_{i^{\prime}, j}$, and concatenated the observational data sampled from $\mE_{n, i}$ and $\mE_{n, j}$ with the same lengths as the interventional data. The correlation value was from these two concatenated expression vectors. When the interventional data was unavailable, the two vectors for correlation were only the observational data $\mE_{n, i}$ and $\mE_{n, j}$. Gene pairs with correlation values larger than 0.1 were considered positive samples.

Features of each gene pair were also retrieved from their expression data. We first row-wise normalized the expression matrix with the z-score normalization, and for the gene pair $\langle\rg_{i}, \rg_{j}\rangle$, we extracted four features: $\overline{\mE_{n, i}}$, $\overline{\mE_{n, j}}$, the average observational expression of $\rg_{i}$ and $\rg_{j}$, and $\overline{\mE_{i^{\prime}, i}}$, $\overline{\mE_{i^{\prime}, j}}$, the average intervened expression by $\rg_{i}$. When the interventional data was unavailable, we filled the last two with 0 and NaN.

Finally, we initialized the LightGBM model with the parameters in Table~\ref{hyperparameters}, and trained and inferred on the entire dataset, which generated the optimal results in our experiments. Sorting the gene pairs descending according to the model inferred scores. For the challenge, the top 1000 pairs as our submission were chosen.

\begin{table}[t]
\caption{LightGBM hyper-parameters}
\label{hyperparameters}
\small
\begin{center}
\begin{tabular}{ll}
\multicolumn{1}{c}{\bf Parameter}  &\multicolumn{1}{c}{\bf Value}
\\ \hline \\
boosting\_type          &gbdt \\
objective               &binary \\
metric                  &binary\_logloss \\
num\_leaves             &5 \\
max\_depth              &2 \\
min\_data\_in\_leaf     &5 \\
learning\_rate          &0.05 \\
min\_gain\_to\_split    &0.01 \\
num\_iterations         &1000 \\
\end{tabular}
\end{center}
\end{table}

\paragraph{SparseRC: Learning Gene Regulatory Networks under Few-Root-Causes assumption} \citep{misiakos2023learning, misiakos2023learning2}
The few-root-causes assumption suggests that the gene activation measurements can be modeled as a linear system whose input is approximately sparse. The underlying gene-gene interaction network is assumed to be a directed acyclic graph (DAG) and the data are generated with a linear structural equation model (SEM) on the DAG. 

Consider a DAG $\mathcal{G}=(V,E)$ with $V=\{1,2,...,d\}$ with adjacency matrix $\mA=(a_{ij})_{i,j\in V}$ where $a_{ij}$ denotes the weight of the edge from $i$ to $j$. Then, we say the data $\mX\in\R^{n\times d}$ follow a linear SEM \citep{shimizu2006lingam,zheng2018dags} if they are generated via the equation
\begin{equation}
    \mX = \mX\mA + \mN \Leftrightarrow \mX\left(\mI - \mA\right) = \mN,
    \label{eq:SEMrecursive}
\end{equation}
where $\mN$ is (typically) assumed to be i.i.d. noise. Since $\mA^d = {\bm 0}$ we have $(\mI - \mA)^{-1} = \mI+\mA+\mA^2+...+\mA^{(d-1)} = \mI+\overline{\mA}$, and Eq. (\ref{eq:SEMrecursive}) can be written equivalently as
\begin{equation}
    \mX = \mN\left(\mI+\overline{\mA}\right).
    \label{eq:SEMcloseform}
\end{equation}
Intuitively, Eq. (\ref{eq:SEMcloseform}) indicates that the linear SEM can be viewed as a process that takes initial input values $\mN$ that we call {\em root causes}, which percolate through the DAG to produce the output $\mX$. This viewpoint provides the motivation to consider an alternative input of values $\mC$ that carry actual information. Adding noise $\mN_C$ on $\mC$ and noise $\mN_{X}$ in the measurement of $\mX$ we obtain the novel formulation
\begin{equation}
    \mX = \left(\mC+\mN_C\right)\left(\mI+\overline{\mA}\right) + \mN_X.
    \label{eq:SEMrootcauses}
\end{equation}
Further, we assume that only a few root causes trigger the output, i.e., that the input $\mC+\mN_C$ is approximately sparse. Interestingly, this assumption can be viewed as a form of Fourier sparsity \cite{bastiJournalpaper,bastiSparseFunctionsOnDAGs}. This setting is identifiable \cite{misiakos2023learning2} and the DAG can be recovered by solving the following optimization problem:
\begin{equation}
\begin{split}
    \min_{\mA \in \R^{d\times d}} \frac{1}{2n}\left\|\mX - \mX\mA\right\|_1 \quad \text{ s.t.}\quad h\left(\mA\right)= 0,
    \label{eq:MobiusDAGopt}
\end{split}
\end{equation}
where $h(\mA)=0$ is the continuous acyclicity constraint pioneered by NOTEARS \cite{zheng2018dags}.

To handle data with interventions included in the competition dataset, a masking matrix $\mM\in \R^{n\times d}$ is used, which captures the intervention on gene $i$ by removing the incoming edges to node $i$. Thus, $\mM$ consists of all ones, except in row $i$, which is set to zero. Since the positions of the interventions in the dataset are known the final optimization problem becomes
\begin{equation}
\begin{split}
    \min_{\mA \in \R^{d\times d}} \frac{1}{2n}\left\|\mX - \mX \mA\odot\mM\right\|_1 \quad\text{ s.t.}\quad h\left(\mA\right)= 0.
    \label{eq:intevenMobiusDAGopt}
\end{split}
\end{equation}
This method is referred to as SparseRC. Its assumptions and the associated optimization (\ref{eq:intevenMobiusDAGopt}) perform decently on the CausalBench challenge. This suggests that having a few root causes may be biologically relevant, which invites further investigation.

\paragraph{Differences in Mean Expression} \citep{kowiel2023causalbench}
Another approach involved directly measuring the strength of the causal relationship $g_i \to g_j$ for every gene pair $(g_i, g_j)$, for which interventions on $g_i$ were available. To achieve this, for a given gene $g_j$, we separately calculated its mean expression on the observational data $\overline{ P^{\emptyset}(\rvx_{g_j})}$ and on the $g_i$-perturbed interventional data $\overline{P^{\sigma(\rvx_{g_i})}(\rvx_{g_j})}$. The absolute difference in means, $|\overline{ P^{\emptyset}(\rvx_{g_j})} - \overline{P^{\sigma(\rvx_{g_i})}(\rvx_{g_j})}|$, was used to measure the strength of the relationship, and then to sort gene pairs. With such a ranked list of genes, we selected the top-$k$ pairs with the largest differences, where $k$ is a user-defined parameter. We note that for MeanDifference estimation, we did not employ any gene expression thresholds.

\paragraph{CATRAN: Causal Transformer} \citep{bakulin2023catran}
CATRAN is a method that uses a transformer-like architecture to learn causal relations between genes. Its core idea is that the degree to which one gene influences another can be represented as a similarity between their learnable vector embeddings. In contrast to the original DCDI approach, this method allows to avoid explicitly learning the adjacency matrix of the gene-regulatory network, which is quadratic to the number of genes. This significantly reduces the number of parameters in the neural network and regularizes its behavior. 

In order to learn good gene embeddings, CATRAN implements a  strategy commonly used for pre-training transformers. Namely, reconstructing the perturbed data. In the case of CATRAN, the perturbation consists of randomly shuffling 80\% of the gene expression values in each training batch. These expression values are then appended to the learnable embeddings of the corresponding genes. Next, to retrieve the original data, these embeddings are iteratively updated by sharing the information between genes: each embedding is replaced with a weighted sum of all others. The weights here are calculated as the sigmoid of the dot product between the original gene embeddings and are identical between iterations. Finally, the expression values are regressed from the updated gene embeddings and the reconstruction loss is calculated.

In addition to reconstruction loss, CATRAN calculates interventional loss which takes advantage of the information on the type of perturbation in each cell. The idea behind it is that for each link between a perturbed gene and other genes, we can estimate its importance by looking at how essential the first gene is to predicting the expression of the second one. This is inspired by \citep{wu2022causally}. The premise is that if one gene is associated with another, then its expression should help the model to make accurate predictions. And so the attention score for each link is estimated using the following formula: 
\begin{equation}
    \sigma\left (\frac{\textrm{Huber}(\textrm{non-interv-pred}, \textrm{non-interv-true})}{\textrm{Huber}(\textrm{interv-pred}, \textrm{interv-true}))} - 1 \right ) .
\end{equation} In the end the interventional loss is estimated using the Huber function \citep{Huber1964} as the error between attention scores and importance scores.


\paragraph{Other proposed methods} 
The other submitted methods mainly looked into fine-tuning the DCDI algorithm. Unfortunately, those attempts did not yield significant improvements, which confirms that new algorithms need to be developed for this task and that the low-performance of existing methods is not due to insufficient hyper-parameter search.

\section{Results and analysis}

\begin{figure}[!h]
    \centering
        \centering
        \includegraphics[width=0.9\textwidth]{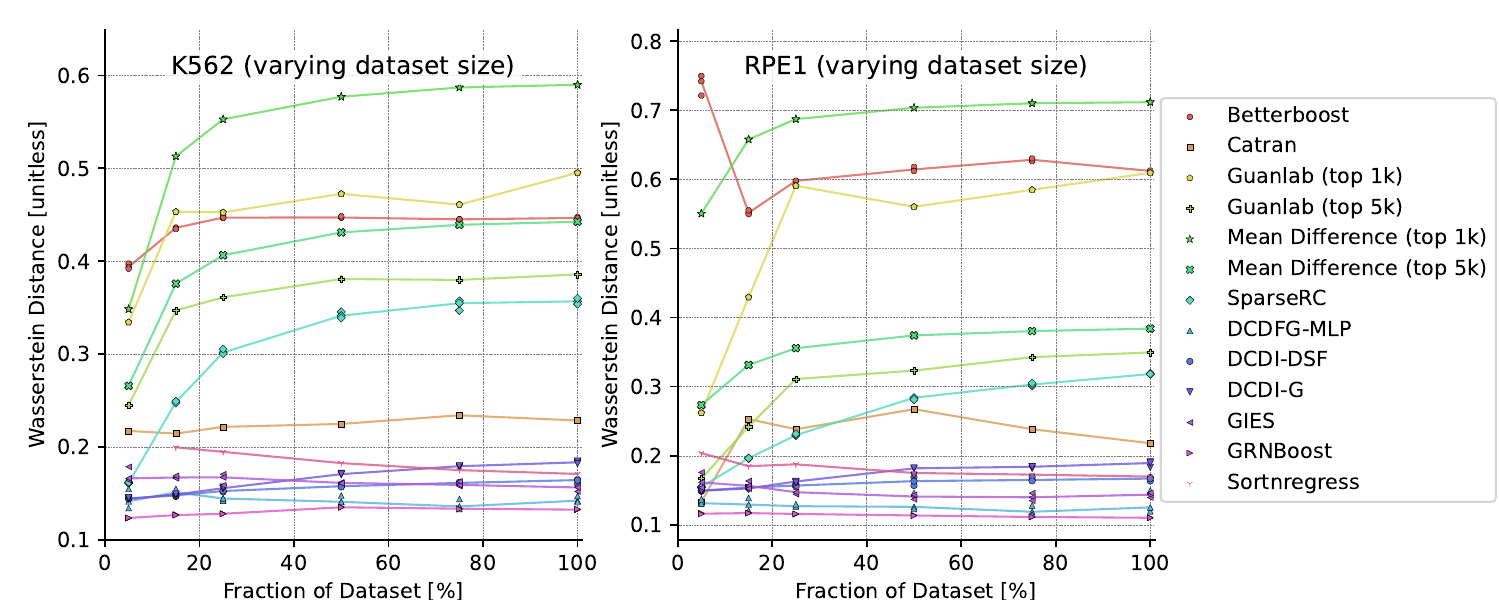}
        \includegraphics[width=0.9\textwidth]{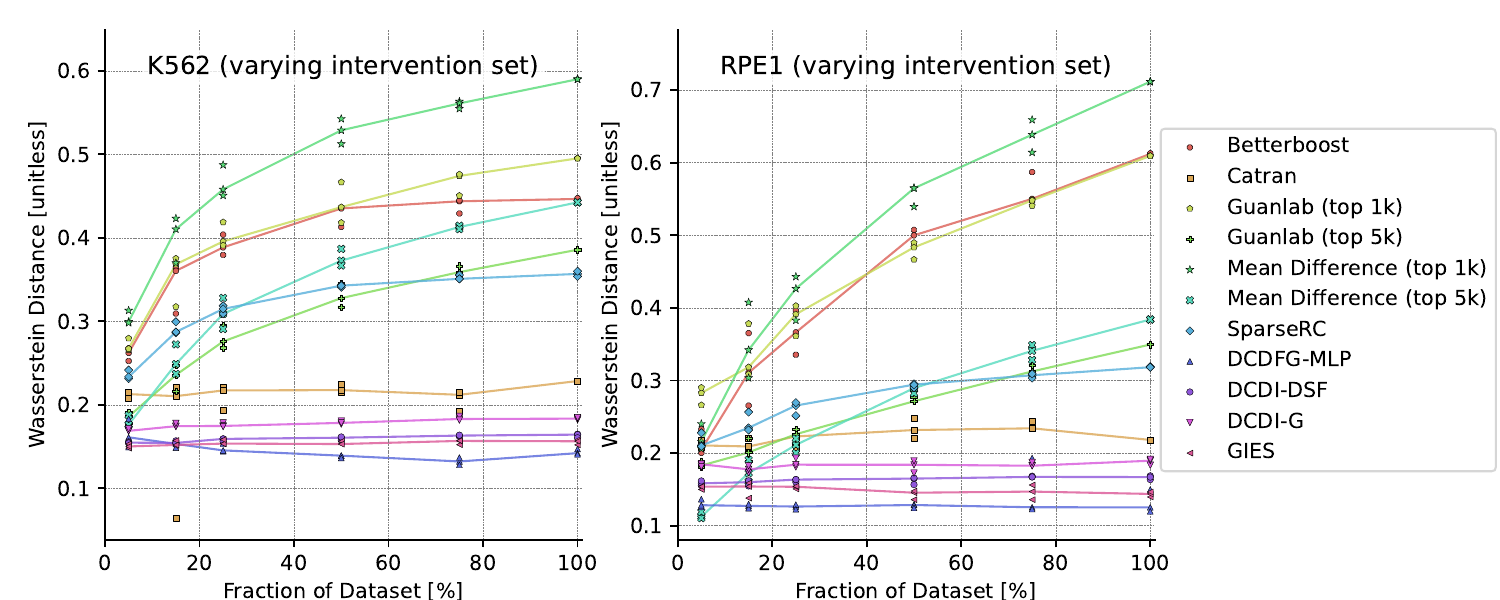}
    \caption{Performance comparison in terms of Mean Wasserstein Distance (unitless; y-axis)  (top row) when varying the fraction of the full dataset size available for inference (in \%; x-axis), and (bottom row) when varying the fraction of the full intervention set used (in \%, x-axis). Markers indicate the values observed when running the respective algorithms with each of three random seeds , and colored lines indicate the median value observed across all tested random seeds for a method.}
    \label{fig:wasserstein_distance_effect}
    
\end{figure}

\begin{figure}[!t]
    \centering 
    \begin{adjustbox}{center, max width=10cm}
    \includegraphics{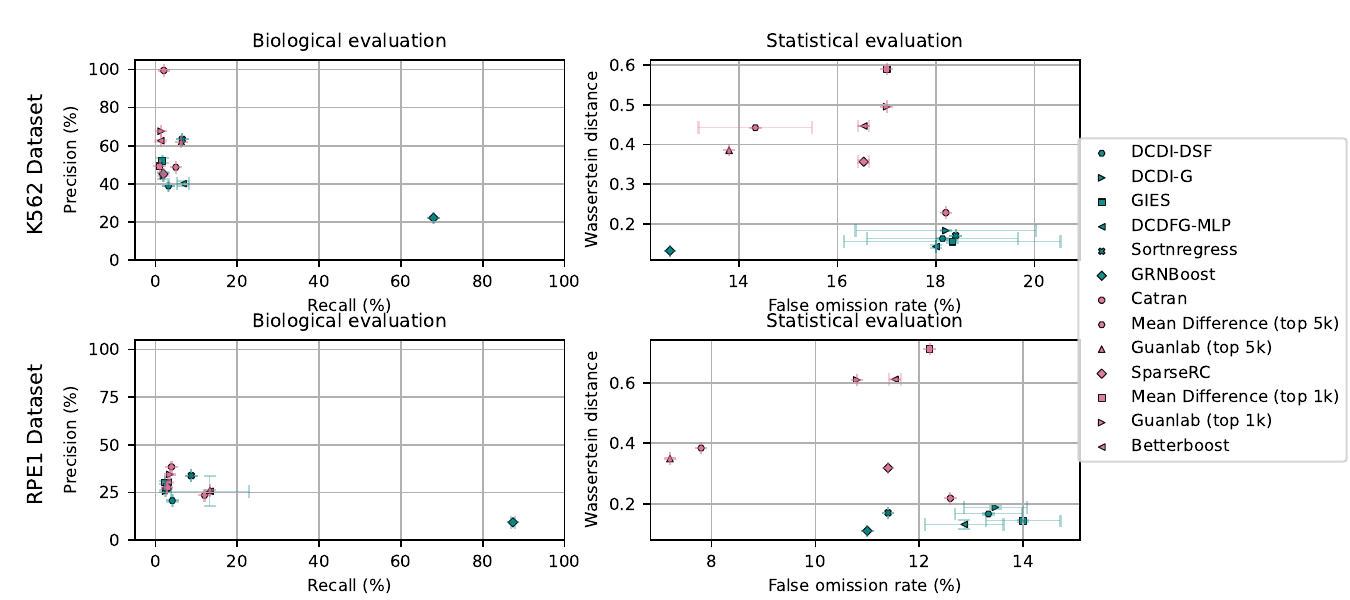}
    \end{adjustbox}
    \caption{Performance comparison in terms of Precision (in \%; y-axis) and Recall (in \%; x-axis) in correctly identifying edges substantiated by biological interaction databases (left panels); and our own statistical evaluation using interventional information in terms of Wasserstein distance and FOR (right panels). For each method, we show the mean and standard deviation from three independent runs. Baseline methods are in green, and the challenge methods are in pink.}
    \label{fig:precision_recall}
\end{figure}

We present here an empirical evaluation of the proposed methods using CausalBench and compare them to some baselines, namely GRNBoost \citep{aibar2017scenic}, DCDI \citep{brouillard2020differentiable}, DCDFG \citep{lopez2022large}, Sortnregress \citep{reisach2021beware} and GIES \citep{hauser2012characterization}. In \cref{fig:wasserstein_distance_effect}, we evaluate the scalability of those methods in terms of the number of samples and the number of interventions. We note here that the evaluation metrics used for the competition were tailored to encourage scaling in terms of the number of interventions. One of the surprising findings of CausalBench \citep{chevalley2022causalbench} was that existing methods showed little or no scaling in terms of interventions, which indicated that interventional methods do not fully leverage the interventional data and that there is an untapped potential to harvest the information provided by interventions. Additionally, in \cref{fig:precision_recall}, we evaluate the trade-off that those methods exhibit in terms of biological and statistical evaluation. In the statistical evaluation, the mean Wasserstein distance serves as the precision, and the false omission rate (FOR) as the recall. We lastly produce a final ranking in \cref{tab:ranking_k562} and \cref{tab:ranking_rpe1} similarly to the one presented in \citet{chevalley2022causalbench}. 

\paragraph{Proposed methods significantly improve utilization of the interventional data} The submitted  methods show significant utilization of the interventional data as can be seen in \cref{fig:wasserstein_distance_effect}. The methods Betterboost, Guanlab, MeanDifference, and SparseRC all exhibit an upward trending performance as more interventional data is provided to those models. This indicates that they leverage the interventional data, which is a major improvement compared to the previous state of the art, where methods showed no benefit from the additional interventional data.

\begin{table}[t]
    \centering
    \small
    \caption{Complete ranking on the K562 cell line. }
\begin{tabular}{lrrrll}
\toprule
           Model &  Rank  &  Rank  &  Mean Position & Wasserstein  & FOR \\
           & Wasserstein & FOR & & Distance & \\
\midrule
Guanlab (top 5k) & 5 & 2 & 3.5 & 0.386 ± 0.000 & 0.138 ± 0.000 \\
MeanDifference (top 5k) & 4 & 3 & 3.5 & 0.442 ± 0.000 & 0.143 ± 0.013 \\
Betterboost & 3 & 4 & 3.5 & 0.447 ± 0.001 & 0.165 ± 0.001 \\
MeanDifference (top 1k) & 1 & 6 & 3.5 & 0.590 ± 0.000 & 0.170 ± 0.000 \\
Guanlab (top 1k) & 2 & 7 & 4.5 & 0.495 ± 0.000 & 0.170 ± 0.000 \\
SparseRC & 6 & 5 & 5.5 & 0.357 ± 0.004 & 0.165 ± 0.001 \\
GRNBoost & 13 & 1 & 7.0 & 0.133 ± 0.000 & 0.126 ± 0.000 \\
Catran & 7 & 10 & 8.5 & 0.228 ± 0.000 & 0.182 ± 0.000 \\
DCDI-DSF & 10 & 9 & 9.5 & 0.163 ± 0.003 & 0.181 ± 0.017 \\
DCDI-G & 8 & 11 & 9.5 & 0.183 ± 0.001 & 0.182 ± 0.021 \\
DCDFG-MLP & 12 & 8 & 10.0 & 0.143 ± 0.004 & 0.180 ± 0.000 \\
Sortnregress & 9 & 13 & 11.0 & 0.171 ± 0.000 & 0.184 ± 0.000 \\
GIES & 11 & 12 & 11.5 & 0.155 ± 0.004 & 0.183 ± 0.025 \\
\bottomrule
\end{tabular}
\label{tab:ranking_k562}
\end{table}

\begin{table}[t]
    \centering
    \small
    \caption{Complete ranking on the RPE1 cell line. }
\begin{tabular}{lrrrll}
\toprule
           Model &  Rank  &  Rank  &  Mean Position & Wasserstein  & FOR \\
           & Wasserstein & FOR & & Distance & \\
\midrule
Guanlab (top 5k) & 5 & 1 & 3.0 & 0.350 ± 0.000 & 0.072 ± 0.000 \\
MeanDifference (top 5k) & 4 & 2 & 3.0 & 0.384 ± 0.000 & 0.078 ± 0.000 \\
Guanlab (top 1k) & 3 & 3 & 3.0 & 0.609 ± 0.000 & 0.108 ± 0.000 \\
Betterboost & 2 & 7 & 4.5 & 0.611 ± 0.002 & 0.115 ± 0.001 \\
MeanDifference (top 1k) & 1 & 8 & 4.5 & 0.712 ± 0.000 & 0.122 ± 0.000 \\
SparseRC & 6 & 5 & 5.5 & 0.318 ± 0.001 & 0.114 ± 0.000 \\
Sortnregress & 9 & 6 & 7.5 & 0.170 ± 0.000 & 0.114 ± 0.000 \\
Catran & 7 & 9 & 8.0 & 0.218 ± 0.000 & 0.126 ± 0.000 \\
GRNBoost & 13 & 4 & 8.5 & 0.110 ± 0.000 & 0.110 ± 0.000 \\
DCDI-G & 8 & 12 & 10.0 & 0.188 ± 0.005 & 0.135 ± 0.007 \\
DCDI-DSF & 10 & 11 & 10.5 & 0.166 ± 0.003 & 0.133 ± 0.007 \\
DCDFG-MLP & 12 & 10 & 11.0 & 0.132 ± 0.018 & 0.129 ± 0.009 \\
GIES & 11 & 13 & 12.0 & 0.143 ± 0.004 & 0 .140 ± 0.008 \\
\bottomrule
\end{tabular}  
    \label{tab:ranking_rpe1}
\end{table}

\paragraph{Proposed methods significantly improve the trade-off between mean Wasserstein and FOR} The proposed methods, especially Guanlab, MeanDifference and Betterboost, offer an advantageous balance between mean Wasserstein and FOR compared to previously existing methods. 
We note that the \emph{precision-recall} frontier, or performance trade-off, of these three methods, is closely aligned, as can be observed in \cref{fig:precision_recall} and when looking at the rankings in \cref{tab:ranking_k562,tab:ranking_rpe1}, where those methods have similar mean positions. This parallel performance isn't surprising given the models' shared assumptions, indicating an improved balance in this \emph{precision-recall}-like trade-off. Lastly, we can note the comparatively higher performance of CATRAN in terms of the biological evaluation, especially on K562. However, CATRAN shows little improvement in terms of intervention scaling. We can thus hypothesize that the gene embeddings learned by CATRAN mainly encode the co-expression of genes. Those links are not necessarily causal, but they nevertheless perform well in terms of biological evaluations as co-expressed genes are often reflected in reference databases \citep{gillis2014bias}. 

\section{Discussion and future work}

\looseness=-1 Despite the significant improvement in performance displayed by the methods submitted to the CausalBench challenge, we must underscore a few limitations, which signal the necessity for continued innovation and progress. One key takeaway from these methods lies in the strategic approach toward identifying potential downstream genes when interventional data is available for a given gene. This involves a direct comparison of the distribution across control and perturbed settings, an insight utilized in different ways by Guanlab, MeanDifference, and Betterboost. Unfortunately, this approach limits their broader applicability, as these methods do not extrapolate well to interactions lacking interventional data. Furthermore, these three methods assess relationships on a case-by-case basis, failing to consider the entire graph, which raises the likelihood that many predicted relationships are not direct but rather mediated causal effects. Thus, further developments are needed for improved differentiation between direct and mediated causal effects. SparseRC stands out as the only method anchored in the principles of causality which offers identifiability guarantees. Despite trailing slightly behind the top-performing methods, SparseRC outperformed other causal baselines considerably. This indicates that theory-grounded causal methods still hold substantial promise. Notably, such methods may be quite valuable in the case where perturbations are not available for all genes (see \cref{tab:ranking_k562_25,tab:ranking_rpe1_25} where SparseRC performs much better comparatively). Furthermore, the inherent assumption of SparseRC, namely \emph{the few-root-causes} assumption, may have biological relevance worthy of further investigation. However, the method's efficiency is hindered by the presumption of linearity in interactions and the equal weighting assigned to all samples, regardless of their observational or interventional nature.
The novel approach adopted by CATRAN, aimed at learning gene embeddings, provides another exciting avenue for exploration. The method's effectiveness could potentially be boosted by better leveraging the available interventional data.

These observations underscore the rich and expansive scope for future research in this domain. With continued innovation and refinement, there is immense potential to unlock further advancements in the field of gene interaction network inference, thereby accelerating our understanding of complex biological systems.

\begin{table}[t]
    \centering
    \small
    \caption{Complete ranking on the K562 cell line with only 25\% of interventional data. }
\begin{tabular}{lrrrll}
\toprule
           Model &  Rank  &  Rank  &  Mean Position & Wasserstein  & FOR \\
           & Wasserstein & FOR & & Distance & \\
\midrule
MeanDifference (top 1k) & 1 & 3 & 2.0 & 0.401 ± 0.032 & 0.177 ± 0.011 \\
SparseRC & 4 & 1 & 2.5 & 0.291 ± 0.008 & 0.164 ± 0.006 \\
MeanDifference (top 5k) & 5 & 2 & 3.5 & 0.253 ± 0.020 & 0.175 ± 0.005 \\
Guanlab (top 1k) & 2 & 7 & 4.5 & 0.354 ± 0.036 & 0.179 ± 0.023 \\
DCDI-G & 7 & 4 & 5.5 & 0.175 ± 0.002 & 0.177 ± 0.005 \\
Betterboost & 3 & 10 & 6.5 & 0.345 ± 0.035 & 0.183 ± 0.022 \\
DCDI-DSF & 9 & 5 & 7.0 & 0.155 ± 0.002 & 0.178 ± 0.004 \\
Guanlab (top 5k) & 6 & 9 & 7.5 & 0.233 ± 0.018 & 0.181 ± 0.016 \\
GIES & 10 & 6 & 8.0 & 0.153 ± 0.004 & 0.179 ± 0.001 \\
DCDFG-MLP & 11 & 8 & 9.5 & 0.152 ± 0.004 & 0.181 ± 0.025 \\
Catran & 8 & 11 & 9.5 & 0.165 ± 0.099 & 0.184 ± 0.019 \\
\bottomrule
\end{tabular}
\label{tab:ranking_k562_25}
\end{table}

\begin{table}[t]
    \centering
    \small
    \caption{Complete ranking on the RPE1 cell line with only 25\% of interventional data. }
\begin{tabular}{lrrrll}
\toprule
           Model &  Rank  &  Rank  &  Mean Position & Wasserstein  & FOR \\
           & Wasserstein & FOR & & Distance & \\
\midrule
MeanDifference (top 1k) & 1 & 1 & 1.0 & 0.351 ± 0.060 & 0.106 ± 0.016 \\
SparseRC & 4 & 2 & 3.0 & 0.241 ± 0.015 & 0.109 ± 0.011 \\
Guanlab (top 1k) & 2 & 4 & 3.0 & 0.335 ± 0.043 & 0.131 ± 0.005 \\
Betterboost & 3 & 7 & 5.0 & 0.314 ± 0.057 & 0.134 ± 0.013 \\
MeanDifference (top 5k) & 8 & 3 & 5.5 & 0.175 ± 0.017 & 0.113 ± 0.016 \\
Guanlab (top 5k) & 6 & 5 & 5.5 & 0.203 ± 0.018 & 0.132 ± 0.007 \\
Catran & 5 & 6 & 5.5 & 0.213 ± 0.008 & 0.133 ± 0.001 \\
DCDI-G & 7 & 9 & 8.0 & 0.178 ± 0.005 & 0.141 ± 0.003 \\
DCDFG-MLP & 11 & 8 & 9.5 & 0.127 ± 0.003 & 0.134 ± 0.002 \\
DCDI-DSF & 9 & 10 & 9.5 & 0.159 ± 0.004 & 0.142 ± 0.005 \\
GIES & 10 & 11 & 10.5 & 0.150 ± 0.012 & 0.148 ± 0.002 \\
\bottomrule
\end{tabular}  
    \label{tab:ranking_rpe1_25}
\end{table}


\looseness=-1 In summary, the CausalBench Challenge (CBC) was designed and implemented with the intention of bridging the domain gap observed in interventional causal discovery methods on synthetic data and real-world data. Our goals centered on inspiring and engaging the broader machine learning community to make advancements in gene interaction network inference based on single-cell perturbation data. We achieved this through the establishment of clear goals, well-planned logistics, suitable metrics, and fair ranking procedures. A computational platform was also set up to ensure robust evaluation of the submitted methods.
As a result of our efforts, the challenge garnered numerous insightful submissions that pushed the boundaries of our understanding and application of causal inference methods. By sharing some of the most innovative and effective approaches in this report, we hope to inspire continued research and development in this critical area of causality and bioinformatics. Most importantly, the analysis of the submissions' performance helped establish a new state of the art for gene network inference from perturbational single-cell data. The performance displayed by the winning methods at CBC2023 surpassed the previous state of the art methods, indicating that the machine learning community is making strides toward better leveraging interventional data for inferring gene interaction networks. This represents a significant milestone in our collective journey to better understand gene interactions and potentially develop new genetically-informed drugs. 

\looseness=-1 Furthermore, we strongly believe that the insights from this challenge do not only apply to the gene network domain, but also hold important lessons for causal inference at large. We envision that the proposed methods will inspire the causality community to develop new causal discovery methods that have higher chances to translate to real-world impact. 
The progress achieved thus far serves as a testament to the potential of the field and should serve as motivation to continue pushing the boundaries of what we can achieve with modern causal machine learning methods in genomics. Moving forward, we hope that the CausalBench Challenge and the results it has yielded will continue to inspire research in this area, with the ultimate aim of accelerating the discovery and development of much-needed medicines.

\bibliography{cbc_iclr2023_conference}

\appendix

\section{Challenge design}
\label{app:design}

The primary task of this challenge revolved around the optimization of models and algorithms for interventional data, specifically in the context of gene-gene interactions. The emergence of advanced, high-throughput techniques to measure genetic perturbation responses at the single-cell level has offered a powerful way to generate evidence for causal gene-gene interactions at scale. Directed graphs, inferred through causal graph methods, can help depict these interactions in a cell-specific manner. They make use of the interventional aspect of perturbed single-cell data, shedding light on disease-relevant molecular targets.

The challenge thus invited the machine-learning community to devise innovative methods for graph inference. The primary goal was to optimize the usage of interventional data. This challenge is driven by the startling observation in the CausalBench benchmark \citep{chevalley2022causalbench} that existing methods do not significantly benefit from larger pools of interventional data. In fact, interventional methods do not outperform observational as well as non-causal methods. 

The challenge was based on the benchmark framework of CausalBench \citep{chevalley2022causalbench}, which is a comprehensive benchmark suite for evaluating network inference methods on single-cell perturbation gene expression data.
 
To evaluate network inference methods, \citet{chevalley2022causalbench} proposes to compute the empirical Wasserstein distance as a measure of distributional change under the effect of the intervention on the parent node of each predicted edge in the output graph, then taking the mean over the scores of all edges. For each edge, $g_i \to g_j$, let $P^{\sigma(\rvx_{g_i})}(\rvx_{g_j})$ be the distribution of the values of node $g_j$ when node $g_i$ is perturbed and  $P^{\emptyset}(\rvx_{g_j})$ be the distribution under no intervention (observational distribution). The score of this edge is computed as the empirical estimate of the Wasserstein distance between these distributions using hold-out samples.

As a facilitating starter, the participants were given two possible routes, using distinct baseline implementations: 

\begin{itemize}
    \item Enhancing GRNBoost \citep{aibar2017scenic} to extend its utilization of interventional data.
    \item Refining the DCDI \citep{brouillard2020differentiable} method, acknowledged as the most effective interventional method, to improve its performance.
\end{itemize}

That said, the submitted methods were not restricted to be derived from the above two baselines, and participants showed great interest in developing new and innovative methods.
We standardized the data, designed the metrics, and developed the baseline code, enabling participants to concentrate on the most intriguing aspect of the challenge which was modeling and algorithmic innovation. Stater code and instructions were provided in the challenge Github repository \url{https://github.com/causalbench/causalbench-starter}.

Submissions were evaluated across different ratios of interventional data (25\%, 50\%, 75\%, and 100\%). A model's final score was calculated as the area under the curve of the mean Wasserstein distance against the ratio of interventional data. The better the performance in terms of using interventional data, the higher the score. To also test that the methods demonstrated an improvement in performance given more interventional data, we computed the difference in mean Wasserstein distance between the 25\% and 100\% ratios of interventional data. This evaluation was conducted on the two datasets included in CausalBench, which correspond to two different biological contexts (cell lines RPE-1 and K562). This thus resulted in four rankings, one per metric per dataset. The final comprehensive ranking consisted of the average rank across the four rankings for each method.

\section{Challenge tasks and timeline}
The challenge took place concurrently with the Machine Learning for Drug Discovery (MLDD) workshop at ICLR 2023. The winners were announced on the workshop day, providing them an opportunity to present their solutions.

The challenge was announced on March 1, 2023, and it consisted of two tasks. In the first phase, participants were given the task of designing and implementing their solutions, with a submission deadline of mid-April. This gave them roughly a month and a half to develop their solutions. The second task involved writing a brief report in the format of an extended abstract to elucidate their solution.

Participants were required to submit a three-page report, supplemented by up to 10 pages of references and additional material. This report had to detail their approach thoroughly and it was to be submitted on the challenge's OpenReview page at \url{https://openreview.net/group?id=GSK.ai/2023/CBC}. 

Each report was reviewed to make sure it is aligned with the guidelines of the challenge before making them public. The current comprehensive report of the challenge together with the uploaded detailed report on each submitted method is hoped to support the reproducibility of the results which we believe is crucial for pushing the boundaries of the current state of the art in graph inference for gene interaction networks.

\section{Challenge submission system and platform}
The challenge submission system front-end was provided by EvalAI, an open-source machine learning platform built and maintained by CloudCV. For the users, EvalAI provided a user-friendly overview of the challenge and a place to check the status of their submissions. EvalAI also offered a useful Python tool that could be used for submitting container images containing entrant code directly from the command line. CBC specifically leveraged EvalAI's \emph{remote evaluation} pipeline, using EvalAI as the front end but performing the actual analysis on a back-end controlled by organizers. All submissions were submitted as Docker images directly to EvalAI, which maintained a repository of container images and a message queue of active submissions throughout the competition. After a submission passed through EvalAI's front end, it was then pulled into the specific CausalBench pipeline that archived and ran the submitted code.

The CBC submission backend was built using Google Cloud Platform services and Slurm, a traditional high-performance computing orchestrator. After pulling a submission container image from the EvalAI registry, entrant code was extracted from the submission image, archived in Cloud Storage, installed into a fresh Python virtual environment, and then run using the scheduler. Upon completion, the results were collated for analysis and also archived. While initially, the plan was to run submissions inside containers running on Google Kubernetes Engine (GKE), the large and sometimes inconsistent memory requirements of some entrant code made a more traditional HPC environment more suitable for this purpose. This process also preempted the emergence of minor problems that had been seen in the past, most notably the accidental submission of ARM processor-based container images rather than more typical x86-based containers. It also allowed for both reproducibility and easy intervention by the organizers if a particular submission suffered notable errors.

\end{document}